\documentclass[lettersize,journal]{IEEEtran}
\usepackage{amsmath,amsfonts}
\usepackage{algorithmic}
\usepackage{algorithm}
\usepackage{array}
\usepackage[caption=true,font=normalsize,labelfont=sf,textfont=sf]{subfig}
\usepackage{textcomp}
\usepackage{stfloats}
\usepackage{url}
\usepackage{verbatim}
\usepackage{graphicx}
\usepackage{cite}
\usepackage{booktabs}
\usepackage{multirow}
\usepackage{bm}
\usepackage{tabularx}
\usepackage{color}
\begin{document}
	\title{Imbalanced Large Graph Learning Framework for FPGA Logic Elements Packing Prediction}
	
	\author{
		Zhixiong Di, ~\IEEEmembership{Member,~IEEE}, 
		Runzhe Tao,
		Lin Chen, 
		Qiang Wu, 
		Yibo Lin, ~\IEEEmembership{Member, ~IEEE}
		\thanks{This work was supported by National Natural Science Foundation
			of China (61504110, 62034007). Corresponding author: Zhixiong Di (dizhixiong2@126.com).
			
			Zhixiong Di, Runzhe Tao, Lin Chen and Qiang Wu are School of Information Science and Technology, Southwest Jiaotong University, Chengdu, China. (e-mail:dizhixiong2@126.com, 825140517@qq.com, mix\_lc@qq.com, wq\_cool@126.com). 
			
			Yibo Lin is with the Center for Energy-Efficient Computing
			and Applications, School of Integrated Circuits, Peking University, Beijing, China. (email:yibolin@pku.edu.cn)
		}
	}
	
	\maketitle
	
	\begin{abstract}
		Packing is a required step in a typical FPGA CAD flow. 
		It has high impacts to the performance of FPGA placement and routing. 
		Early prediction of packing results can guide design optimization and expedite design closure.
		In this work, we propose an imbalanced large graph learning framework, ImLG, for  prediction of whether logic elements will be packed after placement. 
		Specifically, we propose dedicated feature extraction and feature aggregation methods to enhance the node representation learning of circuit graphs. 
		With imbalanced distribution of packed and unpacked logic elements, we further propose techniques such as graph oversampling and mini-batch training for this imbalanced learning task in large circuit graphs.
		Experimental results demonstrate that our framework can improve the F1 score by 42.82\% compared to the most recent Gaussian-based prediction method. 
		Physical design results show that the proposed method can assist the placer in improving routed wirelength by 0.93\% and SLICE occupation by 0.89\%.
	\end{abstract}
	
	\begin{IEEEkeywords}
		FPGA, packing prediction, physical design, graph neural networks, imbalanced graph learning.
	\end{IEEEkeywords}
	
	\section{Introduction}
\begin{figure}[htbp]
		\centering
		\includegraphics[width=\linewidth]{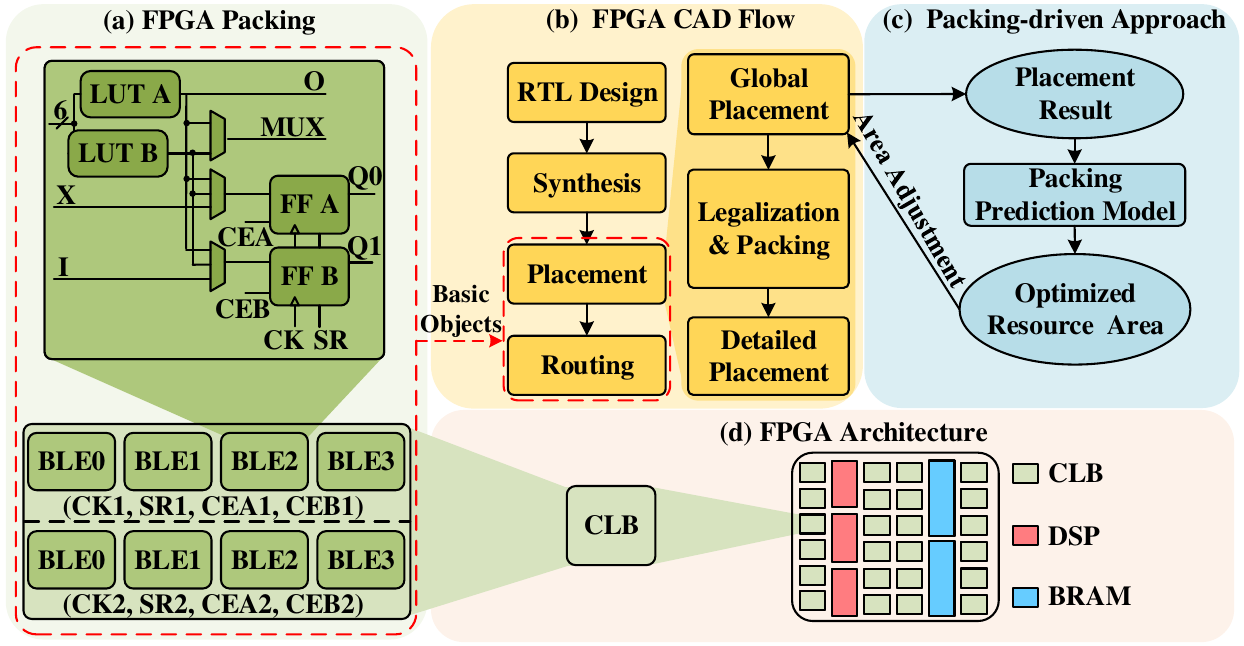}
		\caption{The packing-driven FPGA CAD flow.}
		\label{fig:cad}
	\end{figure}

	\IEEEPARstart{I}{n} a typical FPGA CAD flow, packing clusters all the low-level logic elements such as lookup tables (LUTs) and flip-flops (FFs) into the basic operational objects, i.e., basic logic elements (BLEs) in configurable logic blocks (CLBs) for the placement, as shown in Fig. \ref{fig:cad}. 
	Packing is usually performed during placement and routing (P\&R), and its quality has high impacts on the P\&R closure.
	
	State-of-the-art (SOTA) FPGA placement algorithms \cite{rippleFPGA}--\cite{elfPlace} require packing prediction in the optimization iterations for logic resource estimation. 
	\cite{rippleFPGA} and \cite{Gplace} propose to predict packing by setting static empirical resource demands, which has poor generalization among designs. 
	\cite{DL} and \cite{elfPlace} propose a Gaussian-based approach for packing prediction through estimating resource demands.
	Despite its effectiveness, the previous approaches have two major drawbacks. 
	Firstly, resource demands alone are not enough to decide whether an instance can be packed or not. 
	Secondly, as the number of packed elements is much larger than that of unpacked ones, these approaches are not able to capture the imbalanced distribution of packing results, causing low accuracy on predicting unpacked elements, i.e., high false positive rates. 
	
	As an FPGA design can be represented as a circuit graph, packing prediction can be viewed as a graph learning task on large circuit graphs with physical information and imbalanced label distribution on nodes. Recent studies have shown promising results of leveraging graph neural networks (GNNs) for related tasks like routing congestion prediction in physical design \cite{congnet}~--~\cite{detailroute}. Inspired by these works, in this work, we propose an imbalanced large graph learning framework for packing prediction of logic elements in the FPGA desgin flow. Different from routing congestion prediction, packing prediction is more challenging since the operational objects in this task are instance-level logic elements with complex netlists and imbalanced distribution of packing labels. 
	
	The key contributions are summarized as follows.
	\begin{itemize}
		\item We propose a new graph learning based paradigm for FPGA packing prediction with graph oversampling and mini-batch training to handle imbalanced distribution of packed and unpacked elements.
		\item We propose a region-encoding based feature extraction scheme that aligns with the local nature of packing process.
		\item We propose a homophily-aware feature aggregation method to capture the differences between packed elements and unpacked ones, which enhances the quality of embeddings.
	\end{itemize}
	Detailed experiments demonstrate that our framework outperforms the most recent Gaussian-based method in prediction accuracy. 
	And this technique improves the routed wirelength by 0.93\% and the SLICE occupation by 0.89\% for physical design.
	
	The rest of the paper is organized as follows. 
	Section \ref{preliminaries_and_overview} introduces the background and overview. 
	Section \ref{data_processing} provides the proposed framework. 
	Section \ref{ImLG} demonstrates the results. 
	Section \ref{experiment_results} concludes the paper.
	
	\begin{figure*}[htbp]
		\centering
		\includegraphics[width=\linewidth]{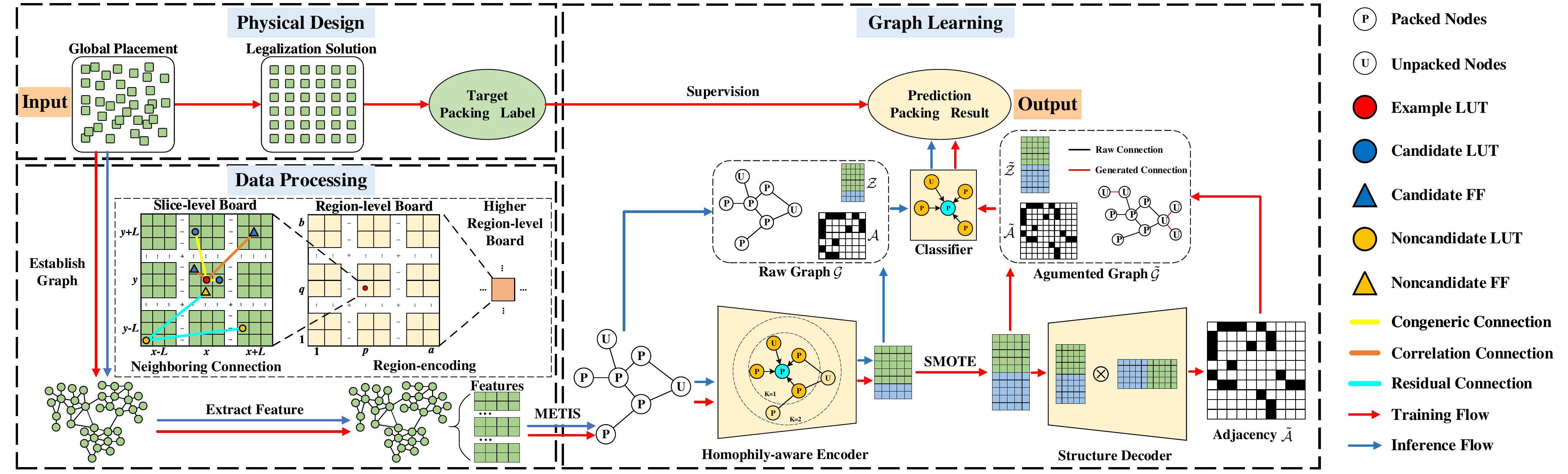}
		\caption{Proposed imbalanced large graph learning framework.}
		\label{fig:framework}
	\end{figure*}
	
	\section{Preliminaries and Overview}
	\label{preliminaries_and_overview}
	\subsection{FPGA Architecture}
	To detail the FPGA architecture, a typical architecture \textit{Xilinx UltraScale} VU095 is illustrated in Fig. \ref{fig:cad}, in which a CLB slice contains eight BLEs and each BLE further contains two LUTs and FFs. 
	The LUTs or FFs satisfying constraints are packed into the high-level BLEs.
	And each CLB is composed of LUTs and FFs with specific logical constraints.
	
	\subsection{Imbalanced Graph Learning for Binary Classification}
	Given the packing prediction can be modeled as a binary node classification task on an imbalanced graph $\mathcal{G}_{im}$, where the number of packed elements is much larger than that of unpacked ones by more than $10:1$. Such biased data greatly weaken the classification performance of the GNN model. The current SOTA methods, such as \cite{imgagn}\cite{smote}, solve this problem with data oversampling algorithms, as it has been found to be the most effective and stable solution. These methods aim to balance the raw data distribution for achieving good classification performance on the augmented balanced graph.
	
	\subsection{Packing Prediction in elfPlace}
	In \textit{elfPlace}, legalization and packing are carried out simultaneously. 
	Therefore, in the global placement stage, \textit{elfPlace} performs a Gaussian method to predict packing results by estimating resource demands and thus assigning appropriate resources for each instance. 
	This packing-driven approach has the ability to enhance circuit quality.
	However, considering the low prediction accuracy of the Gaussian method, a better packing prediction paradigm will be proposed to assist the placer to achieve better solution.
	
	\subsection{Overview of Proposed Framework}
	The overall framework is depicted in Fig. \ref{fig:framework}, which consists of physical design, data processing, and graph learning. The red row represents the training flow, while the blue row represents the inference flow.
	
	The physical design generates the target packing labels through a legalization process. This part is only performed in the training stage. 
	Regarding data processing, the packing-specific connection and encoding scheme are performed to represent a global placement solution as a graph.
	Moreover, graph learning achieves the mapping of nodes to packing results by leveraging the graph oversampling with mini-batch training and the homophily-aware aggregation method.
	With the well-trained GNN model, the packing results of a completely new design can be rapidly inferred without the physical design stage. 
	
	\section{Packing-specific Data Processing}
	\label{data_processing}
	To better training of the GNN model, two packing-specific data-processing methods are proposed, namely, (1) graph establishment with neighbor-priority connection and (2) feature extraction with region-encoding. The first method can effectively reduce the network size and improve the training efficiency, while the second method can extract more informative features for the packing problem.
	
	\subsection{Graph Establishment with Neighbor-priority Connection}
	To establish the graph of the global placement, the difficulty is to build appropriate edges with the fixed number of nodes, since excessive edges bring two defects: (1) high computational burden and (2) redundancy information utilization.
	
	To alleviate these deficiencies, we propose a neighbor-priority connection scheme based on the \textit{Direct-Legalize} \cite{DL} shown in the slice-level board of Fig. \ref{fig:framework}, which are
	(1) \textit{Congeneric Connection}: Node $u$, and its congeneric node $v$ satisfy $max\left(\left|x_v-x_u\right|, \left|y_v-y_u\right|\right) \leq L$, and the packing constraints are connected together, 
	where $(x_{u},y_{u})$ and $(x_{v},y_{v})$ are the coordinates of $u$ and $v$, respectively, and $L$ is empirically set to five CLB slices. (2) \textit{Correlation Connection}: We add connections for LUT and FF, in which the LUT's output pin and the FF's data pin share the same net. This maintains the correlation between the sub-graph of LUTs and FFs. 
	(3) \textit{Residual Connection}: We follow the netlist to build edges between the isolated nodes and other nodes with logic relationships.
	
	The above procedures can greatly reduce the adjacency size and also preserve informative connections as much as possible.
	
	\subsection{Feature Extraction with Region-encoding}
	It is well known that informative features have a very important impact on the performance of machine learning models. However, for packing prediction, two main challenges exist: (1) few attributes can be utilized, and (2) efficient feature representations are desired.
	To overcome these, we propose an attributed feature extraction (AFE) method with a region-encoding scheme. 
	
	In the proposed framework, we make full use of the element type and element location, and encode them as features. First, a six-bit code is used to represent six instance types, which are \{LUT2,$\cdots$, LUT6, FF\}. 
	Second, we adopt a region-encoding scheme to represent the location attribute illustrated in the region-level board of Fig. \ref{fig:framework}. 
	In practice, we partition the layout into multiple regions and then encode the instance's location according to the region in which it falls. 
	By repeating this process, we can customize the location features' dimension.
	This approach can greatly reduce the features' dimension size and generate similar location features for the neighboring logic elements.
	
	\section{Imbalanced Large Graph Neural Networks}
	\label{ImLG}
	We present the imbalanced large graph (ImLG) neural networks, which comprise an improved model based on \cite{graphsmote}. The packing-specific improvements are summarized as follows.
	\begin{itemize}
		\item We propose a homophily-aware aggregation method for the class-rebalanced autoencoder, which captures the differences between nodes.
		\item We add a penalty term for the graph reconstruction error, which enhances the dependability of the augmented graph's topology structure.
		\item We implement cross-graph inductive learning through the mini-batch training strategy.
	\end{itemize}
	
	We first introduce the graph-oversampling-based model architecture shown in Figure \ref{fig:framework}. 
	The raw graph is loaded into the homophily-aware encoder to obtain informative graph embeddings. With the help of synthetic minority oversampling (SMOTE) algorithms \cite{smote}, the minority embeddings are artificially generated. 
	Then, the structure decoder reconstructs the graph topology from the augmented embeddings, and the classifier performs prediction on the augmented graph. 
	With a well-trained encoder and classifier, the prediction labels can be directly inferred on an unseen graph without the SMOTE and graph reconstruction steps.
	
	\subsection{Graph-oversampling-based Model}
	Our model consists of a class-rebalanced autoencoder that implements graph oversampling and a graph-based classifier that enables node binary classification.
	\subsubsection{Homophily-aware Encoder for Embedding Mapping}
	Since the unpacked logic elements are surrounded by the packed ones in the circuit layout, measuring the homophily is an effective way to differ between these two types of elements. 
	Thus, for each node $v$, we implement a specialized aggregation method aiming to capture the homophily, which can be expressed as
	\begin{equation}
		{\mathcal{Z}}_{v} = \sigma\left(\mathbf{W}^{1} \cdot \operatorname{CONCAT}(\mathcal{X}_{v}, \mathcal{X}_{N}^{v}, \mathcal{X}_{N}^{v} -\mathcal{X}_{v})\right),
	\end{equation}
	where $\sigma$ is the activation function, $\mathbf{W}^{1}$ the learnable weight matrix, and $\mathcal{Z}_{v}, \mathcal{X}_{v}$, and $\mathcal{X}_{N}^{v}$ the embedding, raw feature, and aggregated neighbor feature of node $v$, respectively.
	
	\subsubsection{Raw Graph Oversampling}
	We first utilize the SMOTE algorithm to generate the augmented embedding $\widetilde{\mathcal{Z}}$, which is performed by the structure decoder using the inner product to reconstruct the augmented adjacency $\widetilde{\mathcal{A}}$. 
	The adjacency matrix $\widetilde{\mathcal{A}}$ has $\widetilde{\mathcal{A}}_{ij}=1$ if there is an edge between node $i$ and $j$.
	
	\subsubsection{Graph-based Classifier}
	We adopt GraphSAGE for classification on the augmented graph $\widetilde{\mathcal{G}}=(\widetilde{\mathcal{A}}, \widetilde{\mathcal{Z}})$ to output $\widetilde{\mathcal{P}}$, from which the prediction packing labels $\widetilde{\mathcal{Y}}^{\prime}$ can be obtained through an $argmax$ operation.
	\renewcommand{\algorithmicrequire}{ \textbf{Input:}} 
	\renewcommand{\algorithmicensure}{ \textbf{Output:}} 
	\begin{algorithm}[htb] 
		\caption{ Model training algorithm.} 
		\label{alg:model_training_algorithm} 
		\begin{algorithmic}[1]
			\REQUIRE
			Graph  $\mathcal{G}$ with adjacency $\mathcal{A}$, feature $\mathcal{X}$, label $\mathcal{Y}$;
			\ENSURE Optimal network parameters $\bm{\theta}_{Enc}$, $\bm{\theta}_{Dec}$, $\bm{\theta}_{Clf}$ for $Encoder$, $Decoder$, $Classifier$, respectively;
			\STATE $G_{c}$ $\leftarrow$ partition ${G}$ by METIS;
			\STATE $\bm{\theta}_{Enc}$, $\bm{\theta}_{Dec}$, $\bm{\theta}_{Clf}$ $\leftarrow$ initialize network parameters
			
			\FOR{\textit{iter}=$1, \cdots, max\_iter$}
			\STATE $\hat{\mathcal{G}}=(\hat{\mathcal{A}}, \hat{\mathcal{X}}, \hat{\mathcal{Y}})$ $\leftarrow$ random mini-batch from $G_{c}$;
			\STATE $\hat{\mathcal{Z}} \leftarrow Encoder(\hat{\mathcal{A}}, \hat{\mathcal{X}})$;
			\STATE $\widetilde{\mathcal{Z}} \leftarrow$ SMOTE for $\hat{\mathcal{Z}}$;
			\STATE $\widetilde{\mathcal{A}} \leftarrow Decoder(\widetilde{\mathcal{Z}})$;
			\STATE $\widetilde{\mathcal{Y}} \leftarrow Classifier(\widetilde{\mathcal{A}}, \widetilde{\mathcal{Z}})$
			\STATE Compute loss $\mathcal{L}_{rec}, \mathcal{L}_{clf}$ followed Eqs. (\ref{eq:rec}), (\ref{eq:clf});
			\STATE // Update model parameters 
			\STATE $\bm{\theta}_{Enc} \stackrel{+}{\longleftarrow} -\nabla_{\bm{\theta}_{Enc}}(\mathcal{L}_{rec} + \mathcal{L}_{clf})$; 
			\STATE $\bm{\theta}_{Dec} \stackrel{+}{\longleftarrow} -\nabla_{\bm{\theta}_{Dec}}\mathcal{L}_{rec}$;
			\STATE $\bm{\theta}_{Clf} \stackrel{+}{\longleftarrow} -\nabla_{\bm{\theta}_{Clf}}\mathcal{L}_{clf}$;
			\ENDFOR
		\end{algorithmic}
	\end{algorithm}
	
	\begin{table}[htbp]
		\centering
		\caption{ISPD 2016 Contest Benchmarks Statistics}
		\label{tab:ispd2016}%
		\begin{tabular}{|c|c|c|c|c|}
			\hline
			Benchmark & LUT & FF & Minority & Ratio \\
			\hline
			\hline
			FPGA01 & 50K   & 55K   & 8087  &  7.69\%\\
			FPGA02 & 100K  & 66K   & 7969  &  4.57\%\\
			FPGA03 & 250K  & 170K  & 40105 &  9.55\%\\
			FPGA04 & 250K  & 172K  & 46468 &  11.02\%\\
			FPGA05 & 250K  & 174K  & 46773 &  11.02\%\\
			FPGA06 & 350K  & 352K  & 92373 &  13.15\%\\
			FPGA07 & 350K  & 355K  & 96683 &  13.68\%\\
			FPGA08 & 500K  & 216K  & 40055 &  5.59\%\\
			FPGA09 & 500K  & 366K  & 93249 &  10.76\%\\
			FPGA10 & 350K  & 600K  & 107428&  11.31\%\\
			FPGA11 & 480K  & 363K  & 80325 &  9.52\%\\
			FPGA12 & 500K  & 602K  & 86833 &  7.88\%\\
			\hline
		\end{tabular}%
	\end{table}%
	\subsection{Model Optimization with Reconstruction-error Penalty} 
	
	The loss function of the autoencoder can be written as
	\begin{equation}
		\mathcal{L}_{r e c}=\|(\mathbf{A}-\widetilde{\mathbf{A}}_{R}) \odot \boldsymbol{\eta}\|_{F}^{2},
		\label{eq:rec}
	\end{equation}
	where $\widetilde{\mathbf{A}}_{R}$ refers to predicted connections between nodes in the raw graph $G$, $\odot$ represents the Hadamard product, and the penalty term $\boldsymbol{\eta}$ can be written as
	\begin{equation}
		\boldsymbol{\eta}_{i, j}=\left\{\begin{array}{l}
			1 \quad \text { if } \mathbf{A}_{i, j}=0 \\
			\eta \quad \text { otherwise }
		\end{array}\right.,
	\end{equation}
	where $\eta > 1$ imposes more cost on the reconstruction error of the non-zero elements.
	
	The loss function of the classifier is expressed by Eq. (\ref{eq:clf}):
	\begin{equation}
		\mathcal{L}_{clf}=\sum_{u \in \tilde{\mathcal{V}}} \sum_{c\in\mathcal{C}}\left(\mathbf{1}\left(Y_{u}==c\right) \cdot \log \left(\widetilde{\mathcal{P}}_{u,c}\right)\right).
		\label{eq:clf}
	\end{equation}
	where $Y_u$ is the predicted result of node $u$, and $\widetilde{\mathcal{P}}_{u,c}$  the probability that node $u$ belongs to class $c$.
	Above all, the overall optimization objective of the proposed model can be written as
	\begin{equation}
		\min _{\phi, \varphi} \mathcal{L}_{clf}+\lambda \cdot \mathcal{L}_{rec},
		\label{eq:obj}
	\end{equation}
	wherein $\phi$ and $\varphi$ are the parameters for the autoencoder and classifier, respectively, and $\lambda$ is the parameter that controls the trade-off between structure reconstruction and node classification.
	
	\subsection{Graph Partition \& Model Training Algorithm}
	To handle the large placement graph, we partition it into clusters using the graph-clustering algorithm METIS. 
	We set the clustering configuration to create clusters of approximately $10,000$ nodes each, with the \textit{ISPD 2016 contest benchmarks} rounded to the nearest integer to achieve the target cluster size. 
	We use Algorithm \ref{alg:model_training_algorithm} to train our model on these clusters.
	
	\begin{table*}[htbp]
		\centering
		\caption{Comparison of Different Methods on ISPD 2016 Contest Benchmarks}
		\begin{tabular}{|c|cccc|cccc|cccc|}
			\hline
			\multicolumn{1}{|c|}{\multirow{2}{*}{Methods}} & \multicolumn{4}{c|}{\texttt{Gaussian Method} \cite{elfPlace}} & \multicolumn{4}{c|}{\texttt{Cluster-SAGE}\footnotemark[1] \cite{graphsage}} & \multicolumn{4}{c|}{\texttt{Proposed ImLG}} \\
			\multicolumn{1}{|c|}{} &TPR@20 &TPR@40 &F1 score & AUC & TPR@20   & TPR@40 & F1 score & AUC & TPR@20 & TPR@40   & F1 score & AUC \\
			\hline
			\hline
			FPGA01 & 0.2787  & 0.4777  & 0.1368 & 0.5726 & 0.3716 & 0.6276 & 0.4800  & 0.6617  & 0.4746 & 0.7498 & 0.5212  & 0.7254  \\
			FPGA02 & 0.2725 & 0.5450 & 0.0853 & 0.5756 & 0.3201 & 0.6332 & 0.4884  & 0.6488  &  0.4403 & 0.6504 & 0.5122  & 0.7177  \\
			FPGA03 & 0.1783  & 0.3566  & 0.1213 & 0.4885 & 0.4264 & 0.6184 & 0.4804  & 0.6646  & 0.5184 & 0.7336 & 0.5902  & 0.7371  \\
			FPGA04 & 0.1875 & 0.3751 & 0.1436 & 0.4936 & 0.3234 & 0.6504 & 0.4723  & 0.6580  & 0.4907  & 0.7070 & 0.5969  & 0.7199  \\
			FPGA05 & 0.1866  & 0.3731  & 0.1438 & 0.4951 & 0.4029 & 0.6746 & 0.4717  & 0.6827  & 0.4863 & 0.7047 & 0.5932  & 0.7174  \\
			FPGA06 & 0.1807  & 0.3480  & 0.1671 & 0.5281 &  0.3749 & 0.6328 & 0.4921  & 0.6477  & 0.4918 & 0.7285  & 0.5886  & 0.7371  \\
			FPGA07 & 0.1805  & 0.3456  & 0.1688 & 0.5265 & 0.3700 & 0.6730 & 0.4731  & 0.6826  & 0.4789  & 0.6989 & 0.5998  & 0.7115  \\
			FPGA08 & 0.2758  & 0.5515  & 0.1004 & 0.5748 & 0.3413 & 0.6478 & 0.4856  & 0.6843  & 0.4553 & 0.7556 & 0.5188  & 0.7354  \\
			FPGA09 & 0.1988  & 0.3976  & 0.1466 & 0.5145 & 0.3684 & 0.6584 & 0.4715  & 0.6678  & 0.4432 & 0.6985  & 0.5779  & 0.7128  \\
			FPGA10 & 0.1917  & 0.5959  & 0.1526 & 0.6210 & 0.5046 & 0.7902 & 0.4781  & 0.7480  & 0.5968 &  0.8550  & 0.5987  & 0.7789  \\
			FPGA11 & 0.2306  & 0.4576  & 0.1489 & 0.5628 & 0.4527 & 0.7059 & 0.4750  & 0.6885  & 0.5005 & 0.7227 & 0.5351  & 0.7127  \\
			FPGA12 & 0.2799  & 0.4634  & 0.1423 & 0.6198 & 0.5036 & 0.7334 & 0.4995  & 0.7185  & 0.5503 &  0.7787 & 0.5630  & 0.7574  \\
			\hline
			\hline
			Average & 0.2201 & 0.4406 & 0.1381 & 0.5477 & 0.3968 & 0.6705 & 0.4807 & 0.6783 & \textbf{0.4939} & \textbf{0.7320} & \textbf{0.5663} & \textbf{0.7303} \\
			\hline
		\end{tabular}%
		\label{tab:methods}%
	\end{table*}%

	\section{Experiment Results}
	\label{experiment_results}
	\subsection{Experimental Setup}
	To validate the effectiveness of the proposed packing prediction framework, we conduct experiments on \textit{ISPD 2016 contest benchmarks}.
	The detailed statistics of the benchmark are shown in Table \ref{tab:ispd2016}, where ``K'' denotes $1000$, ``Minority'' represents the number of unpacked logic elements, and ``Ratio'' represents the proportion of the minority class.
	
	\footnotetext[1]{An improved version of the Cluster-GCN developed by us.}
    The proposed model is implemented using the PyTorch framework, utilizing a single NVIDIA GeForce RTX 3090 GPU. 
    The Adam optimizer is employed with a learning rate of $1e-3$ and weight decay of $5e-4$ to update the model parameters. 
    The maximum training epoch is set to 1000. The routing process is executed using Vivado tool.
	\begin{table}[htbp]
		\centering
		\caption{Comparison of Effectivenes of Different Feature Extraction Methods}
		\renewcommand\tabcolsep{1.2pt}
		\newcommand{\tabincell}[2]{\begin{tabular}{@{}#1@{}}#2\end{tabular}}
		\begin{tabular}{|c|cc|cc|cc|cc|}
			\hline
			\multicolumn{1}{|c|}{\multirow{2}{*}{Methods}} & \multicolumn{2}{c|}{\texttt{\tabincell{c}{ClusterSAGE\\+GCGE}}} & \multicolumn{2}{c|}{\texttt{\tabincell{c}{Proposed\\ ImLG+GCGE}}} & \multicolumn{2}{c|}{\texttt{\tabincell{c}{ClusterSAGE\\+AFE}}} & \multicolumn{2}{c|}{\texttt{\tabincell{c}{Proposed \\ImLG+AFE}}} \\			
			\multicolumn{1}{|c|}{} & F1 score & AUC   & F1 score & AUC & F1 score & AUC & F1 score & AUC \\
			\hline
			\hline
			FPGA01 & 0.3940  & 0.5242  & 0.4800  & 0.5376  & 0.4800  & 0.6617  & 0.5212  & 0.7254  \\
			FPGA02 & 0.3500  & 0.5422  & 0.4986  & 0.5918  & 0.4884  & 0.6488  & 0.5122  & 0.7177  \\
			FPGA03 & 0.4590  & 0.5303  & 0.4749  & 0.5530  & 0.4804  & 0.6646  & 0.5902  & 0.7371  \\
			FPGA04 & 0.4344  & 0.5192  & 0.4709  & 0.5453  & 0.4723  & 0.6580  & 0.5969  & 0.7199  \\
			FPGA05 & 0.4098  & 0.5409  & 0.4708  & 0.5524  & 0.4717  & 0.6827  & 0.5932  & 0.7174  \\
			FPGA06 & 0.4069  & 0.5292  & 0.4778  & 0.5489  & 0.4921  & 0.6477  & 0.5886  & 0.7371  \\
			FPGA07 & 0.4168  & 0.5120  & 0.4633  & 0.5353  & 0.4731  & 0.6826  & 0.5998  & 0.7115  \\
			FPGA08 & 0.3895  & 0.5077  & 0.4856  & 0.5388  & 0.4856  & 0.6843  & 0.5188  & 0.7354  \\
			FPGA09 & 0.4183  & 0.5216  & 0.4715  & 0.5368  & 0.4715  & 0.6678  & 0.5779  & 0.7128  \\
			FPGA10 & 0.4145 & 0.5233  & 0.4700  & 0.5525  & 0.4781  & 0.7480  & 0.5987  & 0.7789  \\
			FPGA11 & 0.4333 & 0.5088  & 0.4749  & 0.5221  & 0.4750  & 0.6885  & 0.5351  & 0.7127  \\
			FPGA12 & 0.3944 & 0.5122  & 0.4795  & 0.5281  & 0.4995  & 0.7185  & 0.5630  & 0.7574  \\
			\hline
			\hline
			Average & 0.4101 & 0.5226 & 0.4765 & 0.5452 & 0.4807 & 0.6783 & \textbf{0.5663} & \textbf{0.7303} \\
			\hline
		\end{tabular}%
		\label{tab:fe}%
	\end{table}%
	\subsection{Models Evaluation}
	In this subsection, we compare our proposed ImLG with the following two baseline methods.
	\begin{itemize}
		\item \textbf{Gaussian Method}: An analytic packing-prediction approach in \cite{elfPlace} through estimating the resource demands.
		
		\item \textbf{Cluster-SAGE}: An improved version of ClusterGCN \cite{clustergcn} developed by us, which achieves cross-graph training and further improves generalizability.
	\end{itemize}
	
	The evaluation utilizes four standard binary classification metrics: TPR (true positive rate), FPR (false positive rate), AUC (area under the curve), and F1 score. TPR@20 and TPR@40 denote TPR values at FPR=0.2 and 0.4, respectively.
    The experimental results are shown in Table \ref{tab:methods}. 
    Our proposed method outperforms Cluster-SAGE and the Gaussian method by 8.56\% and 42.82\% in F1 score, respectively.

	\begin{figure}[htbp]
		\centering
		\includegraphics[width=0.89\linewidth]{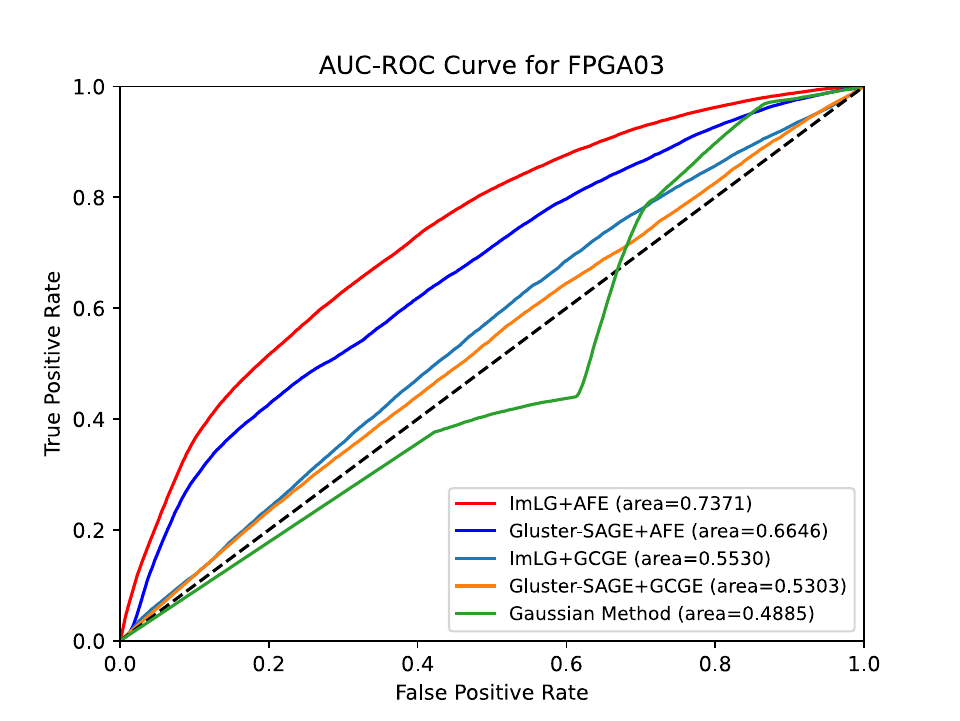}
		\caption{Performance Comparison of Various Models on FPGA03 Using AUC-ROC Curves.}
		\label{fig:methods_comparison}
	\end{figure}
	
	\subsection{Comparison of feature-extraction methods}
    We compare the efficiency of the proposed AFE method and generalizable cross-graph embedding (GCGE) method  \cite{gcge} shown in TABLE \ref{tab:fe}. 
    The column headed ``ImLG+AFE'' indicates that the ImLG model uses the AFE method, and other columns are similar. 
    From the statistics, the proposed AFE method consistently achieves better AUC and F1 score evaluations than the GCGE method, which indicates that the AFE method tends to generate quality features for the packing prediction task. 
    Fig. \ref{fig:methods_comparison} presents a performance comparison of various models on the FPGA03, indicating that graph-based approaches utilizing the AFE method outperform the Gaussian approach by a significant margin.
    
	\subsection{ Improvement on Physical Design}
	We integrate our well-trained packing predictor into our placer and present experimental results in TABLE \ref{tab:pd}. 
	The two columns represent the placement results obtained by the placer using the Gaussian method and the ImLG model to implement the packing prediction, respectively. 
	The``WL'' and ``SO'' metrics indicate the routed wirelength and occupied SLICE sites, respectively. ``WLR'' and ``SOR'' represent the wirelength and occupation ratios normalized to our proposed method.
	The experiments show that our packing predictor improves the routed wirelength by 0.93\% and SLICE occupation by 0.89\%, supporting our assumption that a well-predicted model can guide the placer to achieve better wirelength with minimal resource usage.
    \setlength{\tabcolsep}{1.2pt}
	\begin{table}[htbp]
	\centering
	\caption{Routed Wirelength (WL) and SLICE Occupation (SO) Comparsion on ISPD 2016 Benchmarks}
	\newcommand{\tabincell}[2]{\begin{tabular}{@{}#1@{}}#2\end{tabular}}
	\begin{tabular}{|c|cccc|cccc|}
		\hline
		\multicolumn{1}{|c|}{\multirow{2}{*}{Methods}} &
		\multicolumn{4}{c|}{\texttt{\tabincell{c}{Our Placer\cite{multi-elec} \\+ Gaussian Method\cite{elfPlace}}}} & 
		\multicolumn{4}{c|}{\texttt{Our Placer\cite{multi-elec} + ImLG}} \\
		\multicolumn{1}{|c|}{} & WL & WLR & SO & SOR & WL & WLR & SO & SOR \\
		\hline
		\hline
			FPGA01 & 322575  & 1.0023 & 8189  & 1.0272 & 321843 & 1.0000 & 7972  & 1.0000\\
			FPGA02 & 581221  & 1.0044 & 14722 & 1.0035 & 578955 & 1.0000 & 14671 & 1.0000\\
			FPGA03 & 2875306 & 1.0024 & 37157 & 1.0020 & 2868524& 1.0000 & 37083 & 1.0000\\
			FPGA04 & 4871569 & 1.0016 & 37252 & 1.0020 & 4863815& 1.0000  & 37177 & 1.0000\\
			FPGA05 & 9283586 & 1.0060 & 41197 & 1.0171 & 9228212& 1.0000  & 40504 & 1.0000\\
			FPGA06 & 5729110 & 1.0014 & 55227 & 1.0039 & 5720987& 1.0000  & 55013 & 1.0000\\
			FPGA07 & 8621724 & 1.0039 & 57396 & 1.0119 & 8588057& 1.0000 & 56721 & 1.0000\\
			FPGA08 & 7421319 & 1.0065 & 67057 & 1.0170 & 7373437& 1.0000  & 65933 & 1.0000\\
			FPGA09 & 10587173& 1.0010 & 67145 & 1.0016 & 10576955& 1.0000  & 67035 & 1.0000\\
			FPGA10 & 6131167 & 1.0182 & 65873 & 1.0198 & 6021620& 1.0000  & 64596 & 1.0000\\
			FPGA11 & 10055329& 1.0374  & 67070 & 1.0004 & 9692931& 1.0000  & 67042 & 1.0000\\
			FPGA12 & 6569251 & 1.0271 & 67188  & 1.0003 & 6396026& 1.0000  & 67167 & 1.0000\\
		\hline
		\hline
		Norm. & 6087444 & 1.0093 & 48789 & 1.0089 & 6019255 & 1.0000 & 48409  & 1.0000 \\
		\hline
	\end{tabular}
	\label{tab:pd}
	\end{table}

	\section{Conclusion}
	In this paper, we develop a graph learning based FPGA packing prediction framework that can achieve inductive learning on large FPGA designs with an imbalanced distribution of labels. 
	We present dedicated feature extraction and homophily-aware feature aggregation methods to enhance node representation learning. We further propose techniques like graph oversampling and mini-batching training to tackle imbalanced label distribution in large graphs. 
	Experimental results on the \textit{ISPD 2016 contest benchmarks} showed that the proposed framework outperformed the most recent Gaussian-based method by 42.82\% in F1 score.  
    Physical design results demonstrated that our approach improved routed wirelength by 0.93\% and SLICE occupation by 0.89\%.

\end{document}